\documentclass{homeworg}
\usepackage{times}  % DO NOT CHANGE THIS
\usepackage{helvet}  % DO NOT CHANGE THIS
\usepackage{courier}  % DO NOT CHANGE THIS
\usepackage{graphicx} % DO NOT CHANGE THIS
\usepackage{multirow}
\usepackage{natbib}  % DO NOT CHANGE THIS AND DO NOT ADD ANY OPTIONS TO IT
\usepackage{caption} % DO NOT CHANGE THIS AND DO NOT ADD ANY OPTIONS TO IT
\usepackage{algorithm}

\usepackage[noend]{algpseudocode}
\usepackage{array}
\usepackage{amsmath}
\usepackage{mathtools}
\usepackage{amsthm}
\usepackage{amsfonts}
\newcolumntype{P}[1]{>{\centering\arraybackslash}p{#1}}
\newcolumntype{L}{>{\centering\arraybackslash}m{3cm}}
\usepackage{bm}
\usepackage{subfigure}
\usepackage{geometry}
\usepackage{pdflscape}

\newtheorem{thm}{Theorem}

\newtheorem{defn}[thm]{Definition}

\newcommand{\cN}{\mathcal{N}}

\usepackage{newfloat}
\usepackage{listings}
\DeclareCaptionStyle{ruled}{labelfont=normalfont,labelsep=colon,strut=off} % DO NOT CHANGE THIS
\floatstyle{ruled}
\newfloat{listing}{tb}{lst}{}
\floatname{listing}{Listing}
\AtBeginDocument{%
  \providecommand\BibTeX{{%
    \normalfont B\kern-0.5em{\scshape i\kern-0.25em b}\kern-0.8em\TeX}}}

\begin{document}

% \setcounter{secnumdepth}{0}

\title{Gaussian Process Neural Additive Models}

\author{    Wei Zhang\textsuperscript{\rm 1}, 
    Brian Barr\textsuperscript{\rm 2},
    John Paisley\textsuperscript{\rm 1} \\    
    \textsuperscript{\rm 1}Columbia University, New York, NY, USA\\
    \textsuperscript{\rm 2}Capital One, New York, NY, USA\\
    \{wz2363, jwp2128\}@columbia.edu, brian.barr@capitalone.com}
    \date{}
\maketitle
\section*{Abstract}
Deep neural networks have revolutionized many fields, but their black-box nature also occasionally prevents their wider adoption in fields such as healthcare and finance, where interpretable and explainable models are required. The recent development of Neural Additive Models (NAMs) is a significant step in the direction of interpretable deep learning for tabular datasets. In this paper, we propose a new subclass of NAMs that use a single-layer neural network construction of the Gaussian process via random Fourier features, which we call Gaussian Process Neural Additive Models (GP-NAM). GP-NAMs have the advantage of a convex objective function and number of trainable parameters that grows linearly with feature dimensionality. It suffers no loss in performance compared to deeper NAM approaches because GPs are well-suited for learning complex non-parametric univariate functions. We demonstrate the performance of GP-NAM on several tabular datasets, showing that it achieves comparable or better performance in both classification and regression tasks with a large reduction in the number of parameters.\footnote{This paper appeared at the 2024 AAAI Conference on Artificial Intelligence in Vancouver, BC, Canada.}

\section{Introduction}
With the rapid evolution of deep neural networks, one major challenge has been interpreting and explaining what they learn. DNNs are still generally considered a black-box model because it is difficult to understand and explain why a specific decision is made. This hinders their uptake in some fields such as healthcare and finance, where explainability is highly desired or even mandated by law. While post-hoc explanations can be given using feature importance or counterfactual methods, they do not provide the inherent level of interpretability contained in the weights of a simple linear model, leading some to call for their total rejection in high-stakes problems \cite{rudin2019stop}. 

\begin{figure}[t]
  \centering
  \includegraphics[width=.5\columnwidth]{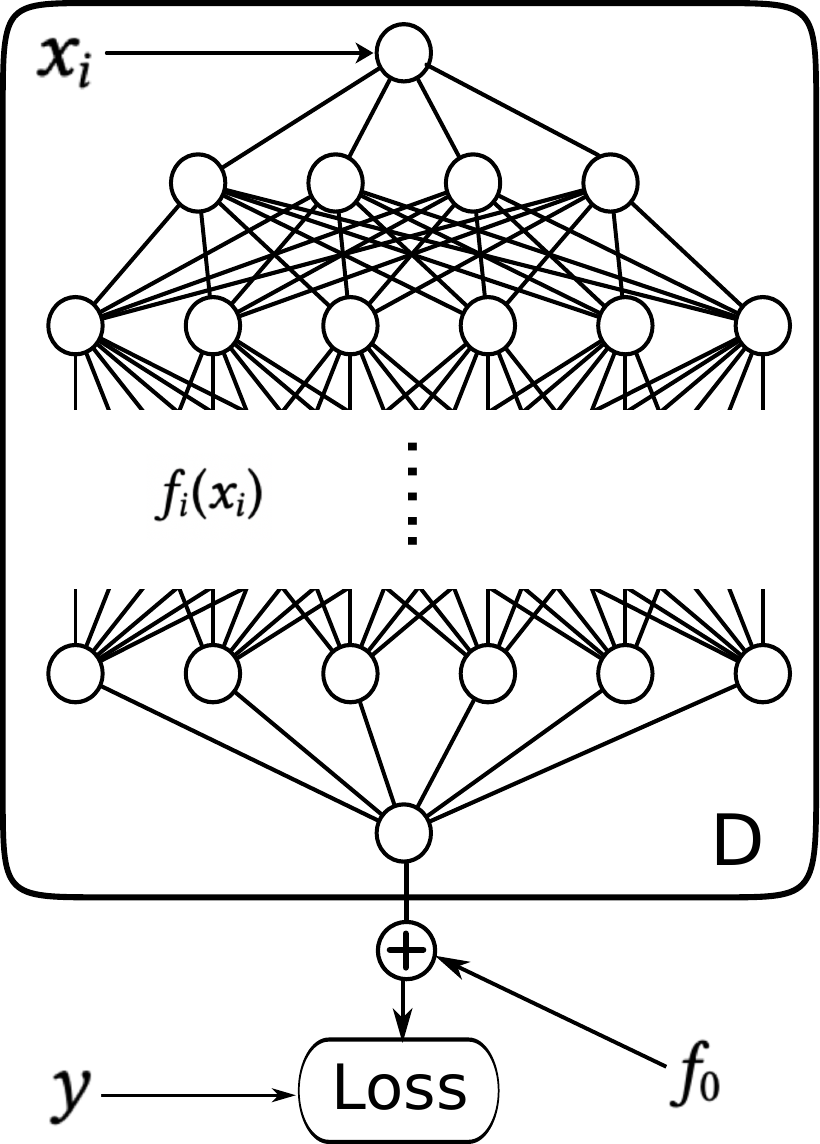}
  \caption{A graphical representation of the neural additive model. $x_i$ is $i$th feature of an input vector having $D$ dimensions. $f_0$ is the bias term. $y$ is the response or label. The function $f_i(x_i)$ is the shape function for feature $i$. The sum $f_0 + \sum_i f_i(x_i)$ is used to predict $y$.}
  \label{fig:nam_arch}
\end{figure}

In this paper, we focus on the family of deep models called neural additive models (NAMs) that attempt to unite the flexibility of deep neural networks with the inherent explainability of linear models. Methodologically, the NAM approximation is formulated as
\begin{equation}\label{eqn.main}
\min_{\theta}~ \mathcal{L}_{\theta}(y,g(x)), \quad g(x) = f_0 + \sum\limits_{i=1}^d f_{\theta_i}(x_i)
\end{equation}
where $x\in \mathbb{R}^d$ is an input vector with $d$ features, $y$ is the target variable and the penalty between $g(x)$ and $y$ can be, e.g., least squares for regression or the logistic regression penalty for classification. See Figure \ref{fig:nam_arch} for an illustration. 

The key innovation of NAM methods is that they learn a feature-specific NN shape function. Whereas linear models simply define $f_{\theta_i}(x_i) = \theta_i x_i$, with $\theta_i$ determining the impact of $x_i$ on $y$, the impact of each feature according to a \textit{nonlinear} function $f_{\theta_i}(x_i)$ can be easily understood by inspection of a one-dimensional plot. Prior to the NAM framework, there have been many candidates for shape functions. For example, \citet{hastie1990generalized} use the spline function. \citet{lou2012intelligible} use boosted decision trees in a method called explainable boosting machines (EBMs). An alternative type of tree called Neural Oblivious Decision Trees \cite{popov2019neural} was proposed as shape function in NODE-GAMs \cite{chang2021node}. Even polynomial regression fits into this framework.

Recently, \citet{agarwal2021neural} proposed using a neural network as the shape function in (\ref{eqn.main}), known as a neural additive model (NAMs). This approach has shown much promise, but comes at the price of potential computational issues. While \citet{radenovic2022neural} do reduce computational expense by sharing neural network layers across features, several practical concerns with neural network training remain. Meanwhile, over the past several decades, researchers have explored the relationship between kernel methods and neural networks \cite{neal2012bayesian}. \citet{cho2009kernel} introduced a new family of positive-definite kernel functions called multilayer kernels, while \citet{lee2017deep} studied infinitely wide neural networks and the Gaussian process, showing the exact equivalence between the two. One of the key findings in \citet{lee2017deep} is that Gaussian process predictions typically outperform those of finite-width neural networks. 

In this paper, we use this GP/DNN insight to propose a novel family of NAMs that leverage the Gaussian process with RBF kernel as the shape function. To this end, we use the random Fourier feature approximation of the Gaussian process \cite{rahimi} and call our framework a Gaussian Process Neural Additive Model (GP-NAM). Unlike other models, the number of parameters in GP-NAM only grows linearly as the input dimension increases, which allows us to train our model quickly, and has a convex objective function which removes dependence on initialization. Furthermore, GP-NAM possesses the same interpretability of related methods since each feature contributes to the output through its own one-dimensional Gaussian process. 

In the next section we review related works. We then review the Gaussian process and random Fourier feature approximation. We then propose our model based on the Gaussian process neural network framework and present an algorithm for learning its parameters. We experiment with several public tabular data sets for regression and classification.

\section{Related Works}
Our work is connected to two research directions: feature importance methods and generalized additive models. 

\paragraph{\textbf{Feature importance.}} Feature importance methods assess the contribution made by each input feature to the output. In linear models this is simply observed by the magnitudes of each learned feature weight (assuming relevant standardization). However, neural networks do not directly provide this information, leading to various post-hoc techniques to analyze the decision boundary. LIME \cite{ribeiro2016should} uses locally linear model approximations around data points to do this, which has consistency issues in the explanations given.  SHAP \cite{lundberg2017unified} is another linear surrogate model-based approach. Integrated gradients (IG) \cite{sundararajan2017axiomatic} and DeepLIFT \cite{shrikumar2017learning} give their explanations by comparing reference points and input points. In contrast to local, data point specific explanations, global attribution methods provide global explanations by clustering vectors pointing from the input data points to the decision boundary \cite{ibrahim2019global}. 

\paragraph{\textbf{Generalized additive models (GAMs).}} GAMs learn the shape function for each individual feature and approximate a target value through a link function \cite{wood2017generalized,hastie1990generalized}. There are many candidate shape functions, such as splines \cite{hastie1990generalized}, random forests \cite{lou2012intelligible} and polynomials \cite{dubey2022scalable}. Recently, deep neural networks have been employed: \citet{agarwal2021neural} construct a 3-layer NN for each input feature independently and introduce a new activation function called ExU for modeling jagged functions. \citet{chang2021node} use neural oblivious decision trees \cite{popov2019neural} as the shape function. \citet{radenovic2022neural} introduce a shared 3-layer NN as a basis function that maps each feature onto a vector space. \citet{bouchiat2023laplace} utilize Bayesian NNs as the shape function and the Laplace approximation to learn its posterior. All these methods are inherently explainable while suffering little loss in prediction performance for many popular tabular data sets. However, they also suffer from scalability issues and usually require regularization techniques such as batch normalization \cite{ioffe2015batch} and dropout \cite{srivastava2014dropout} to prevent them from overfitting. A primary contribution of our proposed GP-NAM framework is the avoidance of these issues. 

\paragraph{\textbf{Additive Gaussian Processes.}} In a parallel line of work, additive GPs have been investigated outside of the neural network modeling framework. \citet{plate1999accuracy} constructs an additive Gaussian Process model and shows the trade-off between interpretability and predictive performance. \citet{duvenaud2011additive} develops further on additive Gaussian Process model and introduces additive kernel. \citet{lu2022additive} discussed the identifiability issue from \citet{duvenaud2011additive} among shape functions. To mitigate this, they use the squared exponential kernel as the base kernel to construct the decomposed kernel functions with constraints. Then, they use Sobol index to measure the contributions of each component to the overall model.

\section{Background: GPs and RFF Linearization} \label{sec_gp}
Before discussing its extension to the NAM regression and classification frameworks, we briefly review Gaussian process (GP) regression and the random Fourier feature (RFF) approximation, highlighting its mathematical equivalence to a single-layer neural network. Given data $(x_1,y_1),\dots,(x_n,y_n),$ where $y\in\mathbb{R}$ and $x\in\mathbb{R}^d$, a GP models this as a function $y(x) : x \rightarrow y$ as follows:

\begin{defn}[Gaussian process]
Given a pairwise kernel function $k(x,x')$ between any two points $x$ and $x'$ in $\mathbb{R}^d$, a Gaussian process is defined to be the random function $y(x) \sim {GP}(0,k(x,x'))$ such that for any $n$ data points, $(y_1,\dots,y_n)$ is Gaussian distributed with $n\times n$ covariance matrix $K_n$ where $K_n(i,j) = k(x_i,x_j)$.
\end{defn}

In this section, $x_i$ indicates the $i$th observation in $\mathbb{R}^d$ and not the $i$th feature. We have defined the mean function of the GP to be $0$ and are interested in Gaussian kernels, or radial basis functions (RBF), of the form
\begin{equation}\label{eq.rbfkernel}
k(x,x') \equiv \exp\left\{-\frac{1}{2b^2}\|x-x'\|^2\right\},
\end{equation}
with parameter $b > 0$. We observe that this kernel function models positive correlations based on proximity in the space of $x$ as defined by the scale parameter $b$; as two points become farther apart, their correlation reduces to zero. In the NAM framework of this paper, each dimension of $x$ will be modeled using a separate GP.

The Gaussian process arises by integrating a linear Gaussian model. That is, let $\phi(x)$ be a mapping of $x$ into another space. If we define the linear regression model
\begin{eqnarray}\label{eqn.linear}
y\,|\,x,w &\sim& \cN(\phi(x)^\top w,\sigma^2),\nonumber\\
w &\sim& \cN(0, I), 
\end{eqnarray}
then the marginal distribution over $n$ observations is
$$y~|~x ~\sim~ \cN(0,\sigma^2 I_n + K_n),$$
where the $n\times n$ kernel matrix $K_n(i,j) = \phi(x_i)^\top\phi(x_j)$. Here, $y$ is represented as a noise-added process, but setting $\sigma^2 = 0$ generates the underlying noise-free GP. 

The power of Gaussian process theory arises when the space $\phi(x)$ is continuous or unknown. This is particularly useful for the RBF kernel, since $\phi(x)$ has a Gaussian form and $\phi(x)^\top\phi(x')$ becomes an integral over $\mathbb{R}^d$, and therefore $w$ needs to be defined over a continuous space in $\mathbb{R}^d$.

The linear representation in Equation (\ref{eqn.linear}) is preferable for scalability to large data sets. For the RBF kernel, since $\phi(x)$ is continuous working in this linear space is impossible. However, approximations can be introduced that seek to construct a finite-dimensional vector $\widehat{\phi}(x)$ such that $\widehat{\phi}(x)^\top\widehat{\phi}(x') \approx \phi(x)^\top\phi(x')$. In this paper we will use the random Fourier feature (RFF) approach, which has the nice property of being mathematically equivalent to a single layer of a fully connected neural network. (We will continue to refer to this approximation as $\phi$.) While originally presented for all shift-invariant kernels by \citet{rahimi}, we focus on the RFF approximation to the RBF kernel. In this case, the RFF method approximates Equation (\ref{eq.rbfkernel}) using a Monte Carlo integral as follows.\newline

\begin{defn}[RFF Approximation]
Let $x\in\mathbb{R}^d$ and define a sample size $S$. Generate vectors $z_s \sim \mathcal{N}(0,I)$ in $\R^d$ and scalars $c_s \sim Unif(0,2\pi)$ independently for $s=1,\dots,S$. For each $x$ define the vector $${\phi}(x) = \sqrt{\frac{2}{S}}\left[\cos\Big(z_1^\top x/b+c_1\Big),\dots,\cos\Big(z_S^\top x/b+c_S\Big)\right]^\top$$
Then ${\phi}(x)^\top{\phi}(x') \approx \exp\{-\frac{1}{2b^2}\|x-x'\|^2\}$ with equality as sample dimensionality $S\rightarrow\infty$.
\end{defn}

Using this representation, we can now return to the underlying linear model of the Gaussian process described in Eqs.\ (\ref{eqn.linear}) by learning $w\in\mathbb{R}^S$, which requires that the same sample set $\{(z_s,c_s)\}$ be shared by all data. Since the approximation to the Gaussian kernel holds, the underlying marginal that this linear model approximates is the desired Gaussian process. The values of $\{(z_s,c_s)\}$ are stored for later predictions. Inspection of Definition 2 shows that the function $\phi(x)$ is mathematically equivalent to a single-layer of a neural network in which the weights $z$ and bias $c$ are not learnable and the nonlinearity used is the cosine function.

\section{Gaussian Process Neural Additive Models}

We next show how a simple application of the RFF approximation to the GP results in a Gaussian Process Neural Additive Model (GP-NAM) for which few parameters need to be learned. While being mathematically equivalent to a NAM approach, GP-NAM results algorithmically in a direct application of a simple linear model, such as logistic regression or least squares linear regression, applied to a pre-determined feature mapping $\phi$. We consider this to be a feature of GP-NAM, since it retains the flexibility and interpretability of other NAM approaches while being fast to learn and avoiding optimization issues with locally optimal solutions because of the convexity of its objective function.

\subsection{Basic Setup}
As mentioned in the introduction, a basic neural additive model is of the form
$$g(x) = f_0 + \sum_{i=1}^d f_{\theta_i}(x_i)$$
where $x_i$ is the $i$th dimension of a vector $x\in\mathbb{R}^d$ and $f_{\theta_i}$ is a neural network that maps $x_i$ to its contribution towards the label/response $y$ using parameters $\theta_i$, which are learned from data. For regression, $g$ typically approximates $y$ using the least squares penalty, while for binary classification $g$ is passed through a sigmoid function.

\begin{figure*}[ht!]
\centering
  \includegraphics[trim={0 0 0 0cm},clip,width=1\textwidth]{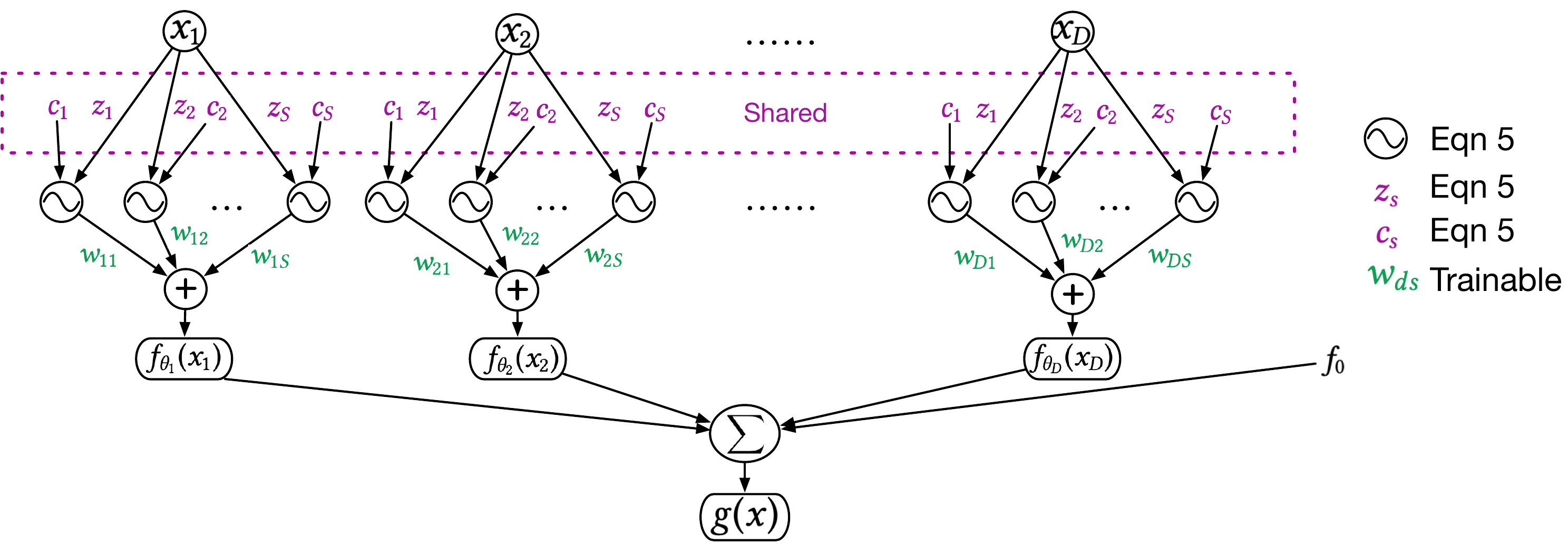}
  \caption{The architecture of GP-NAM. Each $x_i$ represents one feature of a single input vector $x\in\mathbb{R}^d$. Each $(z_s,c_s)$ is shared across shape functions. The GP $f_{\theta_i}(x_i)$ is the shape function for the $i$th feature. The only trainable parameters are the feature-specific $S$-dimensional weight vectors $w_1,\dots,w_D$ that connect the output from the cosine functions to their corresponding GP shape function. The prediction is made by using the sum of the outputs from all the shape functions with the bias term $f_0$. This is mathematically equivalent to an additive Gaussian process.}
  \label{fig:gpnam}
\end{figure*}

In this paper, we define $f$ to be the random function $f_{\theta_i}(x_i) \sim GP(0,k(x_i,x'_i))$ and use the RFF approximation to learn it. Thus the GP-NAM generative process becomes
\begin{eqnarray}
    f_{\theta_i}(x_i) &=& \phi(x_i)^\top w_i,\label{eqn.f}\nonumber\\
    w_i &\sim & \cN(0,I),
\end{eqnarray}
where $\phi(x_i) \in [-\sqrt{2/S},\sqrt{2/S}]^S$ is the RFF map:
\begin{eqnarray}
    \phi(x_i) &=& \sqrt{2/S} \big[\cos(z_s x_i/b_i+c_s)\big]_{s=1}^S,\label{eqn.phi}\nonumber\\
    z_s &\sim & \cN(0,1),\\
    c_s &\sim & \mathrm{Uniform}(0,2\pi).\nonumber    
\end{eqnarray}
We let each dimension of $x$ have its own kernel width $b_i$ to account for different scaling of the features.

We observe that Equation (\ref{eqn.phi}) is a single-layer neural network, while Equation (\ref{eqn.f}) performs the linear operation of a second layer, which is summed over $i$ to either model $y$ directly for regression, or is passed through a sigmoid to model the binary label $y$. (A straightforward extension to multiclass problems can be made, but isn't considered in this paper.) This two-step process is also equivalent to a sum over $d$ dimension-specific Gaussian processes evaluated at each dimension's input. The result is a neural additive model where the only learnable parameters are $w_1,\dots,w_d$, with each $w_i\in \mathbb{R}^S$ and $S$ chosen to provide a good approximation to the GP. Since each GP is one dimensional, we empirically found that $S=100$ works well.

\subsection{Discussion}
As shown in the previous subsection, a GP additive model can be formulated as an equivalent single-layer neural additive model. We provide an illustration of this GP-NAM framework in Figure \ref{fig:gpnam}. In our GP-NAM formulation, only the linear weights of the last layer need to be learned, whereas in the vanilla NAM framework a potentially multilayer neural network is learned for each feature $x_i$ prior to this last linear layer. NBM improves on this by allowing neural network parameters to be shared for each dimension. This leads to some immediate observations.

First, we do not anticipate that GP-NAM will clearly outperform NAM or NBM. This is because, while GP-NAM predefines one layer of parameters $z$ and $c$, other NAM approaches allow these to be learned along with deeper layers, thereby potentially improving the fit. Therefore, since we share $z$ and $c$ across dimensions of $x$, GP-NAM can be considered a special case of NBM that is single layer with unlearnable weights and cosine nonlinearity.

However, in the specific case of additive modeling this is not necessarily a downside, and may be an advantage. While deep neural networks can be expected to outperform Gaussian processes on complex, high-dimensional problems, in the one-dimensional setting of additive modeling it is not clear that a neural network on $\mathbb{R}$ is preferable to a Gaussian process. For one-dimensional function approximation problems a GP with RBF kernel is remarkably flexible in the functions it can learn. Though NBM models can be said to contain GP-NAM as a special case -- just as both can be said to contain the solution set of the classical linear model $x^\top w$ as a special case -- restricting the NBM structure to the GP framework drastically reduces the number of learnable parameters and is convex when $y$ is modeled using least squares or logistic regression. Therefore, learning GP-NAM should be significantly faster and will not suffer from potential local optimal issues arising from the non-convexity of other NAM models.

Finally, we note that extensions to incorporate cross-terms have led to NA$^2$M and NB$^2$M extensions of the form
$$\textstyle g(x) = f_0 + \sum_{i=1}^d f_{\theta_i}(x_i) + \sum_{i' > i} f_{\theta_{ii'}}(x_i,x_{i'}).$$
By letting $\phi(x_i,x_{i'})$ be a two dimensional GP with $z\in\mathbb{R}^2$ as described in the background section, we can extend GP-NAM to GP-NA$^2$M in a similar way. 

\subsection{Algorithm Details}
Since GP-NAM is a linear model applied to a pre-determined feature mapping $\phi$, standard least squares and logistic regression algorithms can be used. We can see the linearity of GP-NAM explicitly by rewriting $g$ as
\begin{equation}\label{eq.linear}
g(x) = w_0 + \sum_{i=1}^d \phi(x_i)^\top w_i % + \sum_{i' > i} \phi(x_i,x_{i'})^\top w_{ii'}
\end{equation}
where $w_0\in\mathbb{R}$ and the remaining $w\in\mathbb{R}^S$ are the only learnable parameters. 
In this section, we provide some additional practical details.

First, we note that the Monte Carlo integral of Definition 2 from which we construct $\phi$ arises from the equality
\begin{equation}
    \exp\Big\{-\frac{1}{2b^2}\|x-x'\|^2\Big\} = \frac{1}{\pi}\int_0^{2\pi}\int_{\mathbb{R}^d}\cos\Big(\frac{z^\top x}{b} + c\Big)\cos\Big(\frac{z^\top x'}{b} + c\Big)\cN(z|0,I)dz dc
\end{equation}
as derived by \citet{rahimi}. For multidimensional problems, being able to sample $(z_s,c_s)$ from a joint Gaussian-uniform distribution greatly simplifies the problem. However, since the NAM framework considers each dimension separately, selecting $z_1,\dots,z_S$ can cbe done using deterministic grid points calculated from the inverse CDF of the standard univariate normal distribution, providing a slightly improved approximation to the integral. For each dimension, we construct $\{(z_s,c_s)\}$ by pairing a randomly permuted uniform grid of $c\in[0,2\pi]$ with a grid of $z\in\mathbb{R}$ using the inverse CDF of $\cN(0,1)$.

For binary classification, the model in (\ref{eq.linear}) can be quickly optimized over $w$ using standard stochastic algorithms for logistic regression. For regularized least squares regression (recalling the Gaussian prior on $w$) the solution to $w$ is closed form, but the dimensionality of the matrix inverse will likely present computational or numerical issues. By stacking $\phi = [1, \phi(x_1),\dots,\phi(x_d)]$ and $w = [w_0, w_1,\dots,w_d]$, the classic conjugate gradients algorithm \cite{nocedal2006} is a fast and stable means for solving
\begin{equation}
\Big(I + \sum_{n=1}^N \phi_n\phi_n^\top\Big) w = \sum_{n=1}^N y_n\phi_n .
\end{equation}
We therefore do not need to invert the left matrix to solve for $w$. We summarize these algorithms in Algorithm \ref{alg}.

\begin{algorithm}[t]
\caption{GP-NAM for regression and classficiation}\label{alg}
\begin{algorithmic}[1]
\Require Data $\{(x,y)\}$, GP width $S$, kernel widths $b_{1:d}$.\\
 \textbf{Sample} $z_s \sim \cN(0,1)$, $c_s \sim \textrm{Unif}(0,2\pi)$, $s = 1:S$.\\ \qquad Alternatively, grid using inverse CDF.\\
 \textbf{Define} $\phi(x_i) = \sqrt{2/S} \big[\cos(z_s x_i/b_i+c_s)\big]_{s=1}^S$\\ \qquad and $\phi_n = \textrm{stack}(1, \phi(x_{1,n}),\dots,\phi(x_{d,n}))$\\
 \textbf{Regression:} Define $A = \sum_n \phi_n\phi_n^\top$ and $v = \sum_n y_n \phi_n$.\\ \qquad Solve $(I+A)w=v$ using conjugate gradients\\
 \textbf{Classification:} Solve linear classifier $w$ on $\{(\phi_n,y_n)\}$\\
 \textbf{Return} $w$
\end{algorithmic}
\end{algorithm}

\section{Experiments}
\begin{figure}[t]  
\centering
\includegraphics[width=.7\columnwidth]{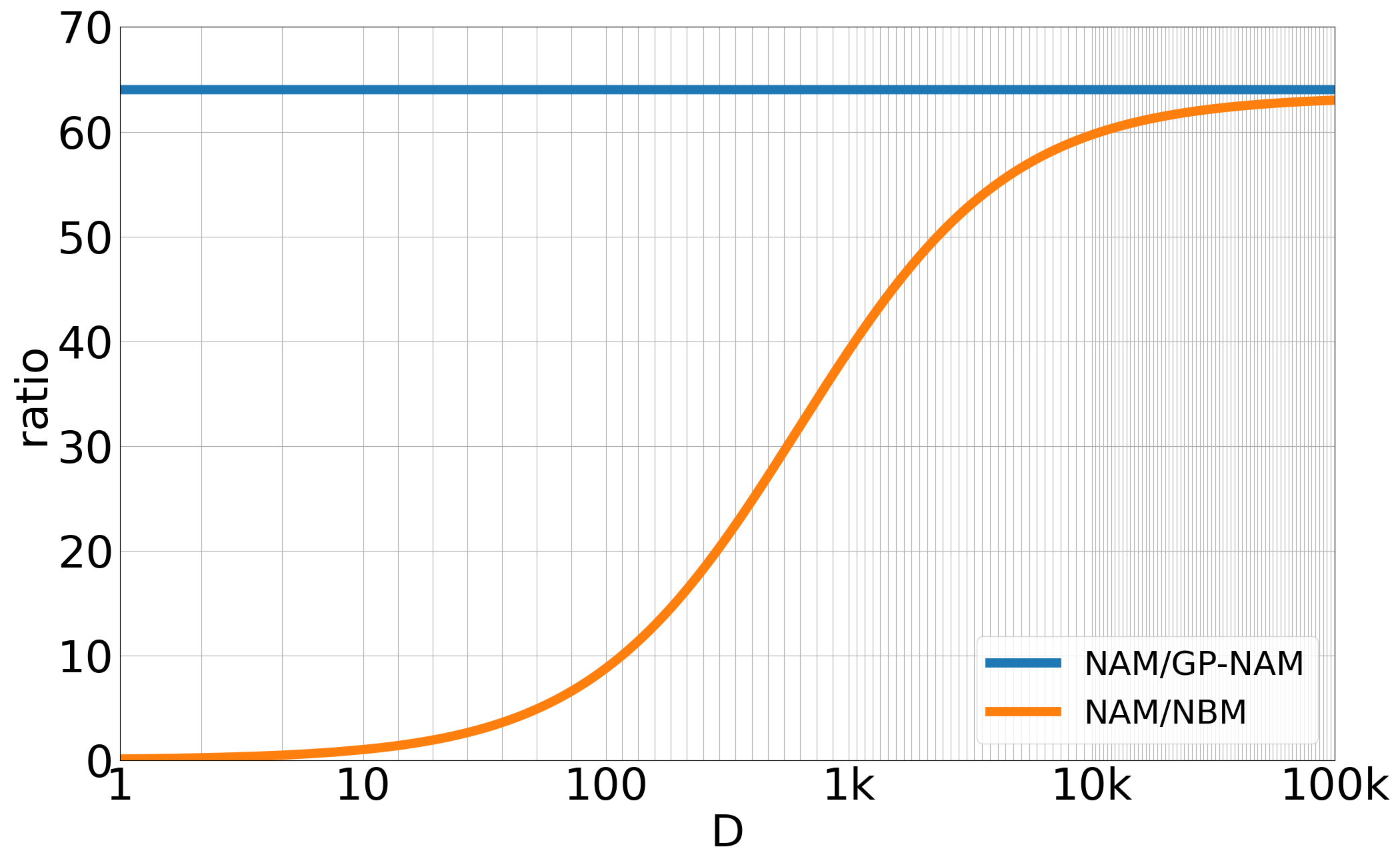}
  \caption{Parameter number ratios $|$NAM$|/|$NBM$|$ (orange) and $|$NAM$|/|$GP-NAM$|$ (blue) as a function of data dimensionality. We set $S=100$ basis functions for all models to give a fair comparison. GP-NAM uses $\sim$60x fewer parameters than NAM regardless of the dimensionality of $x$, and e.g. $\sim$15x fewer than NBM at $x\in\mathbb{R}^{40}$. We are interested in the tabular data regime where e.g. $D<500$.}
  \label{fig:comp_para}
\end{figure}

We experiment using several tabular data sets. A key feature of our model is its reduction in parameters and convex optimization, while still providing competitive performance as a neural additive model. Table \ref{table:runtime} and Figure \ref{fig:comp_para} illustrate the magnitude of this reduction for multiple data scenarios. Here, the parameter size of NAM, NBM and GP-NAM are calculated as $|$NAM$| = (|$NN$|+S)D$, $|$NBM$| = |$NN$| + SD$ and $|$GP-NAM$| = SD$, where $S$ is the basis size, $D$ is the data dimensionality, and $|$NN$|$ is the number of parameters in the neural network. For example, \citet{agarwal2021neural} and \citet{radenovic2022neural} proposed a network of size $|$NN$|=6401$ and $|$NN$| = 62,820$, respectively. The reduction in parameters can lead to a significant improvement in training time. For example, for the LCD data set this translates to 5.5 and 3.5 sec/epoch for NAM and NBM on GPU, respectively, and 50 ms/epoch for GP-NAM on our CPU. For reference, NODE-GAM required 250 ms/epoch and EBM required 50 ms/epoch.

\subsection{Datasets}
We perform experiments on several tabular data sets frequently used for additive regression and classification models. This includes \textbf{CA Housing}\footnote{\url{ https://www.dcc.fc.up.pt/~ltorgo/Regression/cal_housing}}, \textbf{FICO}\footnote{\url{  https://community.fico.com/s/
explainable-machine-learning-challenge}}, for which we follow the processing of \citet{radenovic2022neural}. We also report performance on 
\textbf{MIMIC-II}\footnote{\url{https://archive.physionet.
org/mimic2}}, \textbf{MIMIC-II}\footnote{\url{https://physionet.org/content/mimiciii/}}, \textbf{Credit}\footnote{\url{https://www.kaggle.com/datasets/mlg-ulb/ creditcardfraud}}, \textbf{Click}\footnote{\url{ https://www.kaggle.com/c/kddcup2012-track2}}, \textbf{Microsoft}\footnote{\url{https://www.microsoft.com/en-us/research/
project/mslr}}, \textbf{Year}\footnote{\url{https://archive.ics.uci.edu/ml/datasets/yearpredictionmsd}} and \textbf{Yahoo}\footnote{\url{https://webscope.sandbox.yahoo.com/catalog.
php?datatype=c}}, \textbf{Churn}\footnote{\url{https://www.kaggle.com/blastchar/telco-customer-churn}}, \textbf{Adult}\footnote{\url{https://archive.ics.uci.edu/dataset/2/adult}}, \textbf{Bikeshare}\footnote{\url{https://archive.ics.uci.edu/ml/datasets/bike+
sharing+dataset}} tabular data sets. For these, we follow the processing in \citet{chang2021node,popov2019neural}. We also consider our own processing of credit lending data sets \textbf{LCD}\footnote{\label{github} \url{https://github.com/Wei2624/GPNAM}} and \textbf{GMSC}\footref{github} More information about these data sets is shown in Table \ref{table:datasets}.

\begin{table}[t]
\centering
\begin{tabular}{ c|  c c c c c c   } 
\hline\hline
{\textbf{Model}} &   {\textbf{Bike}} & {\textbf{CA House}}&  {\textbf{FICO}} &  {\textbf{LCD}} \\
\hline
NAM & 54K & 83K   & 262K  & 32K\\
NBM &  65K  & 64K &  68K &   63K\\
GP-NAM & 801 & 1201  & 3901 &  501 \\
\hline
%NA\textsuperscript{2}M &  243K &  & 5.3M & &  \\
%NB\textsuperscript{2}M  & 161K &  & 0.3M & &  \\
%GPNA\textsuperscript{2}M & \\
\hline
\end{tabular}
\caption{Examples of the number of parameters. GP-NAM is a fast, parameter-lite NAM approach with equivalent performance as shown in the quantitative evaluation.} \label{table:runtime}
\end{table} 

\begin{table}[t]
\centering 
\begin{tabular}{ c c c c c  }
 \hline
 \textbf{Dataset} & \textbf{$\#$Train} & \textbf{$\#$Val} & \textbf{$\#$Test} & \textbf{$\#$Feat}\\ 
 \hline
  Churn            & 4,507  & 1,127 & 1,09 & 20  \\
  FICO             & 7,321  & 1,046 & 2,092 & 39 \\
  LCD              & 10,000 & 1,000 &1,000  & 5 \\
  GMSC           & 15,000 & 1,000 &    1,000  & 9     \\
  MIMIC-II        & 17,155 & 2,451 & 4,902 & 17   \\
  MIMIC-III        & 17,502 & 4,376 & 5,470 & 57  \\
  Adult         & 20,838 & 5,210 & 6,513 & 14    \\
  Credit           & 199,364& 28,481&56,962 & 30\\
  Click             & 800,000&100,000&100,000& 11 \\
\hline
  Bikeshare      & 11,122 & 2,781 & 3,476 & 12\\
  CA Housing     & 14,447 & 2,065 & 4,128 & 8 \\
    Year         &370,972 &92,743 & 51,630&90 \\
    Yahoo     & 473,134 & 71,083 & 165,660 & 699\\
  Microsoft         &580,539 &142,873&241,521&136\\
 \hline
\end{tabular}
\caption{Statistics from the tabular data sets used in binary classification (top) and regression (bottom) experiments.}\label{table:datasets}

\end{table} 

\subsection{Baselines}
We compare with several state-of-the-art additive models, as well as two black-box methods for reference. All models are implemented with PyTorch and trained using stochastic gradient descent. 

{\textbf{Linear.}} Logistic and linear regression for classification and regression are the most fundamental and interpretable models. They provide individual weights for each feature.

{\textbf{NAM \cite{agarwal2021neural}.}} NAM is the first model to use neural networks for generalized additive modeling. We use the authors' implementation.

{\textbf{NODE-GAM \cite{chang2021node}.}} NODE-GAM is based on Neural Oblivious Decision Trees (NODE) \cite{popov2019neural} where a full decision tree is learned for each feature. 

{\textbf{NBM \cite{radenovic2022neural}.}} The Neural Basis Model (NBM) reduces the number of trainable parameters with a shared basis neural network that maps each feature to a predefined number of basis features. These are mapped to shape functions by linear projections. 

{\textbf{EBM \cite{lou2013accurate}.}} Explainable Boosting Machines gradient boost thousands of shallow tress for each feature and are considered another type of GAM. We use the interpretML library \cite{nori2019interpretml}. 

{\textbf{MLP.}} Multi-layer perceptrons are a black-box model when interpretability is not needed. We use the architecture reported in \citet{radenovic2022neural}. 

{\textbf{XGBoost \cite{chen2016xgboost}.}} This is another black box baseline. We use the XGBoost library. 

\subsection{Implementation Details}
 In addition to learning shape functions equivalent to a single-layer neural network, the predetermined hidden layer weights and offset means that back-propagation algorithms are unnecessary. Instead, as outlined in Algorithm \ref{alg}, we implement a stochastic optimization algorithm for logistic regression to learn a classifier, or solve the classic conjugate gradients problem for regression, both using the stacked feature mappings $\phi$. Therefore, the algorithm is faster than those available for deeper NAM models.\footnote{The GP-NAM code for regression and classification can be found at \url{https://github.com/Wei2624/GPNAM}}

For NAM, NODE-GAM and NBM, we use the best parameters provided in \citet{radenovic2022neural} or perform a similar hyper-parameter search otherwise. We run Linear, EBMs and XGBoost on CPUs and use the default parameters provided in the libraries. 

We follow other recent papers in calculating AUC or Error Rate to evaluate classification performance, and MSE or RMSE to evaluate regression performance on the same training/validation/testing split for each algorithm. We also provide some qualitative evaluation.

\begin{table*}[t!]
\resizebox{\columnwidth}{!}{
\begin{tabular}{ c| c c c c c c c c c c c}
\hline\hline
\multirow{2}{*}{\textbf{Model}} &\textbf{MIMIC-II}& \textbf{MIMIC-III}&  \textbf{Credit}  &\textbf{GMSC} &\textbf{Adult} &\textbf{Churn} &  \textbf{FICO} &  \textbf{LCD}\\
   &  AUC$\uparrow$ & AUC$\uparrow$ &  AUC$\uparrow$& AUC$\uparrow$ & AUC$\uparrow$ & AUC$\uparrow$ & AUC$\uparrow$ &  AUC$\uparrow$   \\
\hline
Linear & 0.8147 & 0.7753 & 0.9770 &  0.8063 & 0.9013 & 0.8345 & 0.7909 & 0.9459\\
EBM  & 0.8514& 0.8090 & 0.9760    & 0.8655 & 0.9277 & 0.8490 & 0.7985 & 0.9519\\
NAM  & 0.8539& 0.8015 & 0.9766 & 0.8548  & 0.9152 & 0.8356 & 0.7993 & 0.9494 \\
NODE-GAM  &  0.8320 & 0.8140 & 0.9810 & 0.8215 & 0.9166  & 0.8339 & 0.8063 & 0.9558  \\
NBM  &  0.8549 & 0.8120 & 0.9829 & 0.8328 & 0.9176 & 0.8389 & 0.8048 & 0.9506\\
GP-NAM  & 0.8508 & 0.8159 &  0.9794 & 0.8674 & 0.9167 & 0.8360 & 0.8043 & 0.9524 \\
\hline
\hline
\hline
\multirow{2}{*}{\textbf{Model}} &  \textbf{Bikeshare}&    \textbf{Click} &  \textbf{Microsoft} & \textbf{Yahoo} & \textbf{Year}  &  \textbf{CA Housing} && \textbf{Model Size} \\
  & RMSE$\downarrow$ & ERR$\downarrow$ & MSE$\downarrow$ & MSE$\downarrow$ & MSE$\downarrow$  &  RMSE$\downarrow$  && Params$/$feature \\
\hline
Linear & 145.9& 0.3443 & 0.8693  & 0.6765 & 88.51 & 0.7354 & & 1 \\
EBM & 100.0 & 0.3338  & 0.8654  & 0.6312  &  85.15 &  0.5586  & & 10K Stumps\\
NAM & 99.6  & 0.3317 & 0.8588 & 0.6458 & 85.25 &  0.5721  && S$+|$NN$|$\\
NODE-GAM & 100.7  & 0.3342 & 0.8533 & 0.6305  & 85.09 & 0.5658 & & X NODE Tree \\
NBM & 99.4  & 0.3312  & 0.8602 & 0.6384 & 85.10 &  0.5638 && S$+|$NN$|/$D\\
GP-NAM & 99.6 &  0.3030 & 0.8588 & 0.6302 & 85.10 & 0.5586 && S\\
\hline
\hline
\end{tabular}
}
\caption{Quantitative results for several regression and classification tasks. The top half of the plot indicate problems where $\uparrow$ is better, and the bottom half where $\downarrow$ is better. For several data sets, complex additive models improve significantly over the baseline linear model. Among those, GP-NAM performs at the level of more complex alternatives, and occasionally better.} \label{table:stateofart}
\end{table*}

\begin{figure*}[t!]
\centering
  \includegraphics[width=1\textwidth]{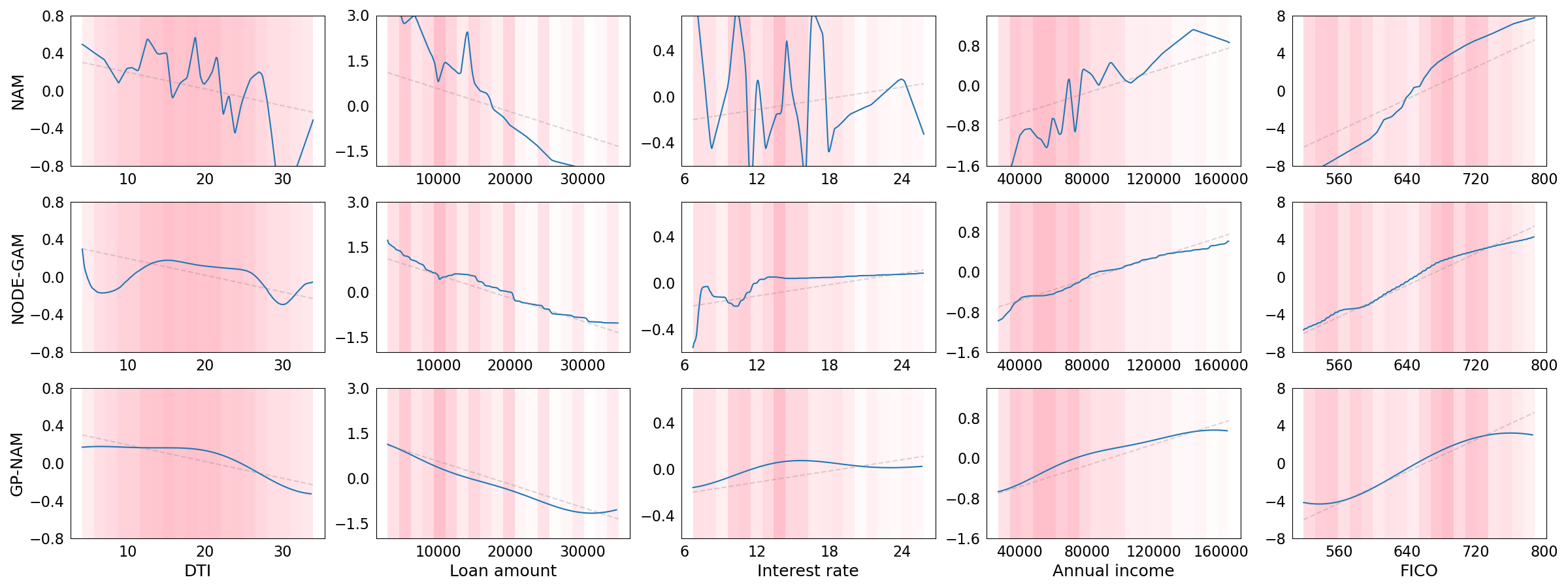}
  \caption{Shape functions of NAM, NODE-GAM and GP-NAM on the LCD data set in the original scales. The density of each feature in the training data is plotted in pink. For reference, logistic regression learned weights indicated by the slope of the light dashed line on each plot. Inspection shows that GP-NAM is in fairly close agreement with linear classification, while still allowing for meaningful nonlinearities to be learned from the data (DTI in particular).}
  \label{fig:shapefunctions}
\end{figure*}

\begin{table}[b]
\centering
\begin{tabular}{ c|  c c c c   } 
\hline\hline
\multirow{2}{*}{\textbf{Model}} &  \textbf{CA House}& \textbf{Bike}& \textbf{FICO} &  \textbf{LCD}\\
  &  RMSE$\downarrow$ & RMSE$\downarrow$ & AUC$\uparrow$ &  AUC$\uparrow$   \\
\hline
Linear & 0.7354 & 145.9 & 0.7909 & 0.9459 \\
GP-NAM & 0.5586 & 99.6 & 0.8043 & 0.9524\\
XGBoost & 0.4428 & 50.0 & 0.7925 &  0.9567 \\
MLP &  0.5041 & 44.2 & 0.7936 & 0.9589\\
\hline
\hline
\end{tabular}
\caption{Performance comparison on a subset of data sets using complex and non-interpretable XGBoost and MLP. Gives an indication of GP-NAM performance in relation to "upper" and "lower" bounds.} \label{table:baseline}
\end{table}

\subsection{Experimental Results and Discussion}

\paragraph{\textbf{Quantitative comparisons.}}
We show our main quantitative results in Table \ref{table:stateofart}. The top half shows results for which higher values are better, and in the bottom half lower values are better. One takeaway from this table is that no NAM approach is clearly best for all problems. GP-NAM occasionally performs best (MIMIC-II, GMSC, Click), or effectively tied for best (Year, CA Housing, Bikeshare, Yahoo, Microsoft). For nearly all of the remaining 6 problems GP-NAM is close to the best along with several other algorithms. Furthermore, for several problems (Credit, Adult, Churn, FICO, LCD, Microsoft) a basic linear model in the features space performs very competitively, indicating a highly linear problem there. For the data sets with obvious improvement (Bikeshare, CA Housing) or moderate improvement (MIMIC-II, MIMIC-III, GMSC, Yahoo), GP-NAM captures the nonlinearity of the problem as well as other NAM approaches. However, as previously highlighted in Table \ref{table:runtime}, GP-NAM is a parameter-light model with a fast convex optimization algorithm. In Table \ref{table:baseline} we also show how GP-NAM can perform well on tabular data in comparison with non-additive models that capture greater complexity in the data, but are harder to interpret. 

\paragraph{\textbf{Interpretability and stability.}} The inherent interpretability of a GAM model can be obtained by visualizing the shape function for each feature. In Figure \ref{fig:shapefunctions} we show the shape functions learned by GP-NAM on the LCD data along with NAM and NODE-GAM, where NAM represents a neural additive model and NODE-GAM a tree-based GAM model. 

As is evident, the neural network of NAM does not learn functions with the same smoothness as the GP -- indeed, it is not clear how interpretable NAM actually is in this case. NODE-GAM is smoother and analysis can determine which, if either, is more meaningful between it and GP-NAM. We note that for GP-NAM, DTI (debt-to-income) follows a meaningful nonlinear pattern, where defaults are consistently lower probability until a $20\%$ threshold is reached, at which point defaults increase in probability with increasing prior debt. In terms of stability under multiple reruns, since GP-NAM is a convex optimization problem there is no randomness on the learned shape functions. The other algorithms required a run to be chosen.

\section{Conclusion}
We have presented a Gaussian Process Neural Additive Model (GP-NAM) for interpretable nonlinear modeling of tabular data. We are motivated by the fact that Gaussian processes are a robust and flexible nonparametric method for univariate function approximation, and thus are as suitable for the generalized additive modeling problem as deeper neural networks. Using the RFF approximation, we demonstrated how GP-NAM is a neural additive model with a single, pre-determined hidden layer and few learnable parameters. The result is an efficient convex optimization problem for regression or classification that performs as well as more complicated, non-convex deep approaches to the problem. Indeed, for low dimensional function approximations, the equivalence of a GP using RFFs with a single layer NN may support a preference for this simpler model over deeper models for certain applications, such as spatio-temporal model averaging \cite{paisley2022bayesian}.

\bibliographystyle{ACM-Reference-Format}
\bibliography{sample-base}

\end{document}